\documentclass{article}

\PassOptionsToPackage{square, numbers}{natbib}

\usepackage[final]{neurips_2024}

\usepackage[utf8]{inputenc} 
\usepackage[T1]{fontenc}    
\usepackage{hyperref}       
\usepackage{url}            
\usepackage{booktabs}       
\usepackage{amsfonts}       
\usepackage{nicefrac}       
\usepackage{microtype}      
\usepackage{xcolor}         

\usepackage{amsfonts}
\usepackage{amsmath}
\usepackage{amssymb}
\usepackage{mathtools}
\usepackage{amsthm}

\title{Revealing the Learning Process in Reinforcement Learning Agents Through Attention-Oriented Metrics}

\author{%
  Charlotte Beylier \\
  Max Planck Institute for Human Cognitive and Brain Sciences\\
  Center for Scalable Data Analytics and Artificial Intelligence (ScaDS.AI) \\
  Dresden/Leipzig, Germany\\
  \texttt{beylier@cbs.mpg.de} \\
  \And
  Simon M. Hofmann \\
  Max Planck Institute for Human Cognitive and Brain Sciences \\
  Leipzig, Germany \\
  \texttt{simon.hofmann@cbs.mpg.de} \\
  \AND
  Nico Scherf \\
  Max Planck Institute for Human Cognitive and Brain Sciences \\
  Center for Scalable Data Analytics and Artificial Intelligence (ScaDS.AI)\\
  Dresden/Leipzig, Germany \\
  \texttt{nscherf@cbs.mpg.de} \\
}

\begin{document}

\maketitle

\begin{abstract}
  The learning process of a reinforcement learning (RL) agent remains poorly understood beyond the mathematical formulation of its learning algorithm. To address this gap, we introduce attention-oriented metrics (ATOMs) to investigate the development of an RL agent's attention during training. In a controlled experiment, we tested ATOMs on three variations of a Pong game, each designed to teach the agent distinct behaviours, complemented by a behavioural assessment. ATOMs successfully delineate the attention patterns of an agent trained on each game variation, and that these differences in attention patterns translate into differences in the agent's behaviour. Through continuous monitoring of ATOMs during training, we observed that the agent's attention developed in phases, and that these phases were consistent across game variations. Overall, we believe that ATOM could help improve our understanding of the learning processes of RL agents and better understand the relationship between attention and learning.
\end{abstract}

\section{Introduction}

Understanding the learning process of a deep Reinforcement Learning (RL) agent is critical to improve its transparency and performance. Yet, the information available during the learning phase of an agent is limited, typically reduced to its performance score. This calls for new metrics to provide insight into how and what an agent learns during training.

Currently, the primary indicator of an agent's learning progress is its performance score derived from task-specific rewards. While necessary to optimize an agent \citep{watkins1992q,konda1999actor,mnih2013playing}, this performance score provides a very limited view of an agent's development during learning and fails to explain or predict certain behaviours in the testing phase \citep{clark2016faulty,cobbe2019quantifying}. In this context, explainable AI (XAI) methods can complement the performance score by providing an insight into an agent understanding of its environment and task through the explanation of its decisions  \citep{verma2018programmatically,shu2017hierarchical}. However, the explanation format of these methods - which may be a decision tree \citep{liu2019toward}, a structural causal model \citep{madumal2020explainable}, or a set of linguistic rules \citep{hein2017particle} -  is often too complex to be tracked during learning. Nevertheless, some studies based on saliency maps \citep{zeiler2014visualizing,zhou2016learning,selvaraju2017grad,simonyan2013deep} have examined changes in an agent's attention during training. For instance, \citep{greydanus2018visualizing} visualized saliency maps of agents trained on Atari games at various stages during training and  \citep{guo2021machine} compared the saliency maps of RL agents and humans playing Atari games during learning. However, these analyses were qualitative and human-driven, resulting in a very sparse source of information during training (fewer than a dozen data samples over the whole training phase). The most closely related work to our study is \cite{lapuschkin2019unmasking} which examined the emergence of a DQN agent's strategy to target the tunnel in Breakout by saving the network's state at various point during training. However, this approach requires waiting until the end of the training to get information about the agent’s attention and is memory intensive. 
Lastly, the need for rigorous scientific methods based on the verification of hypotheses through controlled experiments has grown in the field of explainable reinforcement learning. This was emphasized by \citep{atrey2019exploratory} who used perturbation experiments to test common hypotheses about an agent's strategy when playing ATARI games. Therefore we believe that the new metrics proposed to understand the learning of a RL agents should be accompanied by an appropriate experimental framework and controllable setting.

In this work, we introduce attention-oriented metrics (ATOMs) to gain insight into an agent's learning process through the development of its attention. ATOMs are derived from saliency maps and quantify an agent's attention on the objects within its environment. Specifically, ATOMs encompass two levels of description: the hierarchical-attention (ranked attention on individual objects) and the combinatorial-attention (attention on combinations of such objects). To systematically evaluate ATOMs' ability to give information about what an RL agent has learnt, we created three variations of a Pong game. Each variation required the agent to learn a distinct behaviour. In addition to these games, we designed a behavioural experiment to test if the agent's attention described by ATOMs translated into their observed behaviour \citep{atrey2019exploratory}. 

\section{Method}

\paragraph{Experimental setup}

Data were recorded in actor-critic agents (A2C; \citep{konda1999actor,mnih2016asynchronous}) from the Stable Baseline 3 repository \citep{stable-baselines3}. We study the actor network of each agent, responsible for the choice of action. This network is composed of three convolutional layers followed by a linear layer that we will refer to as \(F_c\) and a final output layer. Throughout the remainder of this paper, we will use the terms 'agent' and 'actor network' synonymously. ATOMs were computed from the neurons in the linear layer (\( F_c \)). This layer is located just before the output layer and encompasses the final \textit{world-model} on which the agent choose its action. Details regarding agent training protocols are provided in appendix \textbf{B}

\paragraph{Attention-Oriented Metrics (ATOMs)}

\begin{figure}
    \centering
    \includegraphics[width=1\linewidth]{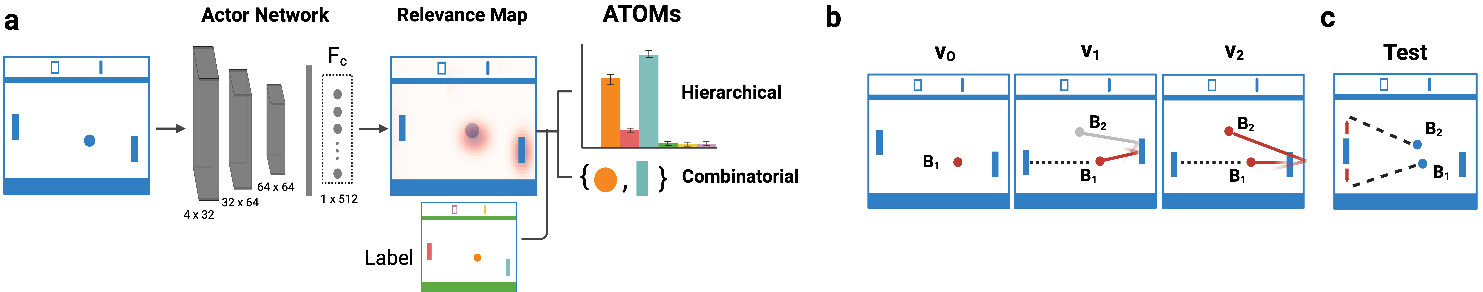}
    \caption{\textbf{a} ATOMs pipeline. \textbf{b} Variations of the Pong game. Red balls yield reward to the agent, the gray ball is a distractor. The balls' dynamics are illustrated with lines indicating if a ball bounces back from the opponent or passes through it. \textbf{c} Dual Ball Discrimination Test, which forces the agent to choose between \textit{B1} or \textit{B2}.}
    \label{fig:atoms}
\end{figure}

Here we characterise the Pong game by its constituent objects: the paddles of the agent and its opponent,  their respective displayed score, the ball(s) (\textit{B1} for \textit{v0}, \textit{B1} and \textit{B2} for \textit{v1} and \textit{v2}) and the walls. Figure \ref{fig:atoms} \textbf{a}  illustrates the process to extract these two metrics. Both metrics that compose ATOMs are derived from the attention of neurons in the \(F_c\) layer that are relevant to an action. Specifically, this attention is measured by computing the Layer-Wise Relevance Propagation (LRP; \citep{bach2015pixel,lapuschkin2019unmasking}). To generate input images \(\pmb{x}\) (consisting of pixels \textit{p}) used for the LRP analysis, we let an agent play the game for ten episodes. Subsequently, we filtered the input to ensure that all objects are present and do not overlap. We then sampled a total of 150 frames and automatically labelled the objects within each frame. Let \(\pmb{X} \in \mathbb{R}^{150\times 4 \times 84 \times 84}\) and \(\pmb{\bar{X}} \in \mathbb{R}^{150\times 4 \times 84 \times 84}\) be the original and the labelled versions of the input set, respectively. We focus on the attention of neurons that influence action choices, identifying relevant neurons in the \(F_c\) layer using LRP. For each input, \(\pmb{x} \in \pmb{X}\) we first calculate the relevance scores \(R_{F_c} = \{R_i\}_{i \in F_c}\) and select neurons \(S \subseteq F_c\) that account for 90\% of the total relevance \(R_{F_c}\). This filtering step is used to avoid noise components and decrease ATOMs computational time. We then generate relevance maps at the model input \(\pmb{R}^k(\pmb{x}) = \{R_p^k(\pmb{x})\}\) for each neuron \(k \in S\). Details regarding the derivation of the relevance scores are provided in appendix \textbf{A}.

\subparagraph{Hierarchical-attention}
For a given object \(o_g \in O\), the hierarchical-attention metric \(h : O \rightarrow \mathbb{R}\) is defined as follows: \(h(o_g) =  \frac{1}{|\pmb{X}|} \sum_{\pmb{x} \in \pmb{X}} \sum_{k \in S}R_k(\pmb{x}) \cdot \bar{R}_{\text{g}}^k(\pmb{x})\) with \(\bar{R}_{\text{g}}^k(\pmb{x}) = \frac{1}{V} \sum_{p \in P_{o_g}} R_p^k(\pmb{x})\) and \(V = |R_{p_g}^k(\pmb{x}) \setminus \{0\}|\) and \(p_g\) denotes all pixels \(p \in o_g\).

\subparagraph{Combinatorial-attention} 
For a subset of objects \(T \subseteq O\), the combinatorial-attention metric \(c : O \rightarrow \mathbb{R}\) is defined as follows: \(c(T) = \frac{1}{|\pmb{X}|} \sum_{\pmb{x} \in \pmb{X}} \sum_{k \in S} R_k(\pmb{x}) \cdot \delta_k(\pmb{x};T)\) 
with \(\delta_k(\pmb{x};T) = 
    \begin{cases} 
        1 & \text{if }  \bar{R}_{\text{g}}^k(\pmb{x}) > \beta \text{ for all } g \in T \\
        & \text{ and }  \bar{R}_{\text{g}}^k(\pmb{x}) = 0 \text{ for all } g \in O \setminus T, \\
        0 & \text{else.}
    \end{cases}\)

where \(\beta = \alpha \cdot M^k\)  with \(\alpha = 0.25\) and \(M^k = \max_{g}(\{\bar{R}_{\text{g}}^k(\pmb{x})\}_g)\).

\subparagraph{Designing Pong variations}

We implemented three different variations of the Pong game (see Figure \ref{fig:atoms} \textbf{b}): \textit{v0}, \textit{v1}, and \textit{v2}. While \textit{v0} maintains traditional gameplay, a second ball \textit{B2}) is introduced in \textit{v1} as a distraction (it does not bring reward). In \textit{v2}, both \textit{B1} and \textit{B2} bring points, but \textit{B2} only interacts with the agent's paddle, awarding +1 for a rebound and -1 if it passes the paddle.Details about the game implementation and the rules can be found in appendix \textbf{C}.

\subparagraph{Dual Ball Discrimination Test}\label{dualtest}
A major difference between the three variations of the game is the importance of ball \textit{B1} and ball \textit{B2}. As the latter only yield rewards in \textit{v2} we expect the behaviour of the agents toward \textit{B2} to change between games. To quantitatively assess agent behaviour, we implemented the \textit{Dual Ball Discrimination Test}, where the agent is placed in a situation that forces a choice between both balls. This experiment involved generating 100 unique trajectories from various initial positions and velocities, designed such that both balls reach the agent's x-coordinate simultaneously while maintaining a separation exceeding the length of the paddle. Consequently, the agent can interact with only one ball, and we can analyse any systematic preference for \textit{B1} or \textit{B2}. An illustration of the test is shown in Figure \ref{fig:atoms} \textbf{c}.

\section{Results}

\paragraph{Evaluation of ATOMs}
We evaluated ATOMs on 50 fully trained agents for each game variation to test ATOMs ability to provide relevant information on the agent's behavior.

\begin{figure*}
     \centering
     \includegraphics[width=1\linewidth]{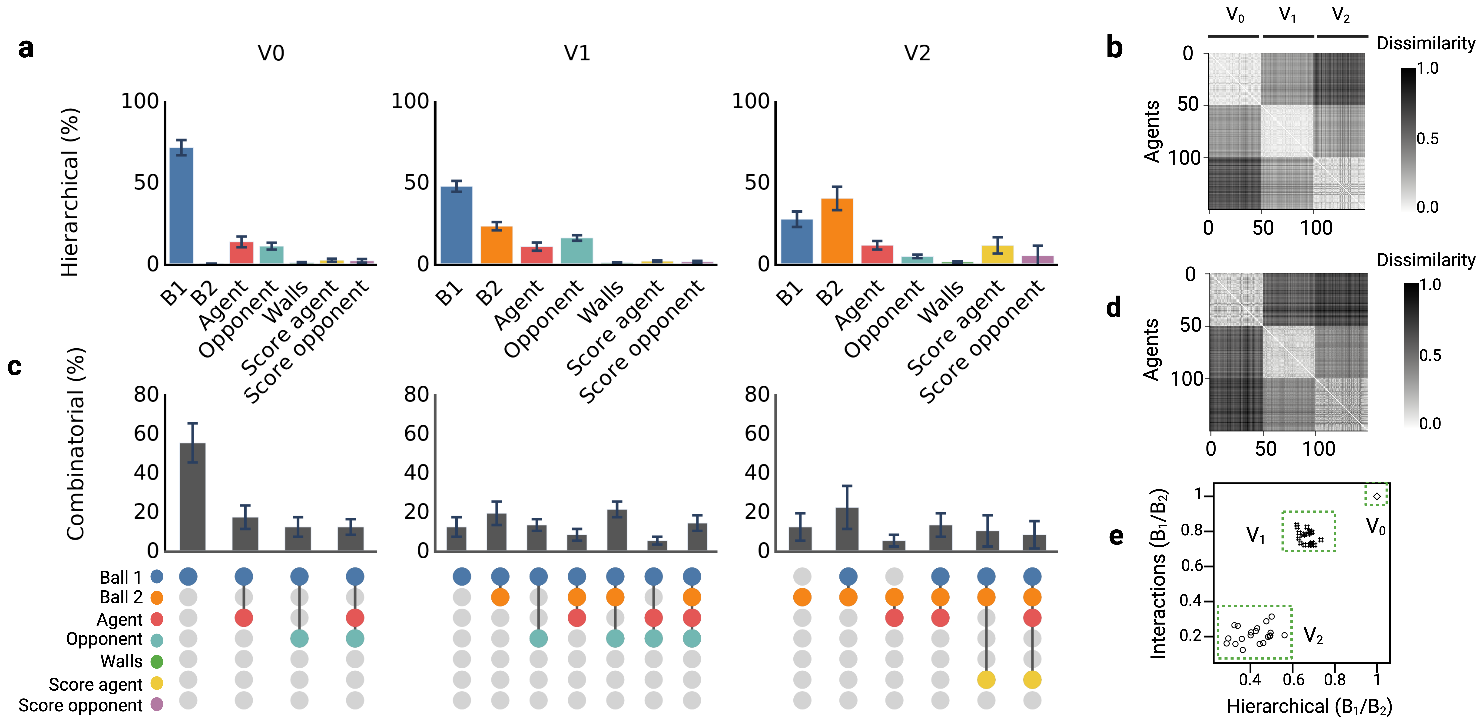}
     \caption{\textbf{a} Hierarchical-metric averaged over all agents. \textbf{b} Dissimilarity matrix for the hierarchy-attention metrics for all agents.\textbf{c} Combinatorial-attention metric which examines co-observation of objects. Each object is symbolized by a distinct coloured dot. Combinations of objects are indicated through the simultaneous colouring of mutliple dots. \textbf{d} Dissimilarity matrix for the combinatorial-attention metrics for all agents. \textbf{e} Relative interaction of the agent with \textit{B1} with respect to \textit{B2} in function of the relative hierarchy of \textit{B1} with respect to \textit{B2}. \textit{v0} was added as a reference.}
     \label{fig:Figure2}
 \end{figure*}

\subparagraph{Different game variations induce game-specific attention pattern.} The hierarchical-attention (Figure \ref{fig:Figure2} \textbf{a}) shows that trained agents attended to the expected aspects of the game: agents trained on \textit{v0} and \textit{v1} paid more attention to the only ball bringing a reward \textit{B1}  while agents trained on \textit{v2} showed a slight preference for the ball \textit{B2}. The combinatorial-attention (Figure \ref{fig:Figure2} \textbf{c}) shows a consistent common pattern across game variations with attention to either the ball(s) alone or to a combination of ball(s) and the agent's paddle or ball(s) and the opponent's paddle. Figures \ref{fig:Figure2} \textbf{b, d} show that successfully trained agents developed similar attention patterns.

\subparagraph{Attention patterns are consistent with agents' behaviour.}

As shown in Figure \ref{fig:Figure2} \textbf{a}, ATOMs indicate that agents trained on \textit{v1} still paid attention to \textit{B2} suggesting that these agents had not learned to ignore the ball completely. We therefore assessed whether increased attention to one ball over the other reflected a preference for interacting with that particular ball using the \textit{ Dual Ball Discrimination Test}. Figure \ref{fig:Figure2} \textbf{e} shows that the relative attention given to each ball indeed reflects how an agent behaves towards them (in terms of preferred interaction). Furthermore, the cluster associated with \textit{v1} shows that these agents still interact with \textit{B2}, confirming that these agents had not learned to ignore \textit{B2} completely. 

\paragraph{Evolution of agents' attention during learning}

\begin{figure*}
    \centering
    \includegraphics[width=1\linewidth]{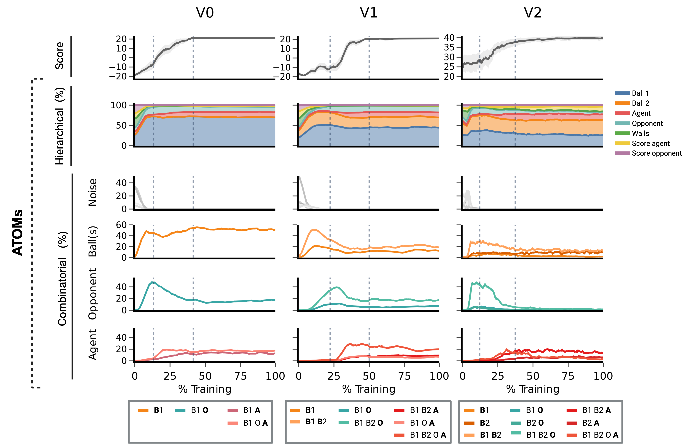}
    \caption{ Performance score and ATOMs during learning.  The agent's performance score was averaged (dark line) over ten games at each measurement (the standard deviation is represented with the shaded region). Dotted lines mark periods of notable score improvement, selected manually. The combinatorial-attention grouped by category: noise (grey), combinations of balls only (orange), combinations of opponent and balls (cyan),and combinations of objects including the agent (red).}
    \label{fig:ATOMS_learning}
\end{figure*}

We then monitored the development of the agents' attention during their training using ATOMs (Figure \ref{fig:ATOMS_learning}). Applied to three variations of a Pong game, we found that the learning process in A2C agents was characterised by a common developmental pattern of attention in successfully trained agents with a late emergence of attention on the agent's paddle concomitant with an increase in its performance score. 

\section{Conclusion}
We propose ATOMS to further our understanding of the learning process of an RL agent's by monitoring the development of its attention. We believe that ATOMs could help explain and test the origin of certain limitations observed in RL agents such as observational overfitting \cite{song2019observational}. In addition, ATOMs could be used as an exploration tool to better understand the relationship between attention and learning in both human \cite{leong2017dynamic} and artificial RL agents. Indeed, convolution neural networks (CNN) have been used to model biological vision \cite{lindsay2021convolutional,choi2023dual}, while reinforcement learning has been used as a framework to study the neurobiology of learning and decision-making \cite{lee2012neural}. Here, one could explore the role of external feedback and TD error (or internal feedback) \cite{jones2010integrating,canas2010attention}, the impact of certain hyperparameters on an agent's attention \cite{guo2021machine} or the interaction between top-down and bottom-up attention during learning.

\section*{Acknowledgments}
N.S., C.B. and S.M.H. are supported by BMBF (Federal Ministry of Education and Research) through ACONITE (01IS22065) and the Center for Scalable Data Analytics and Artificial Intelligence (ScaDS.AI.) Dresden/Leipzig. C.B. is also supported by the Max Planck IMPRS NeuroCom Doctoral Program.
The figures were created using BioRender.com.

\newpage
\appendix


\section{LRP relevance score derivation}

Let \(f : \mathbb{R}^n \rightarrow \mathbb{R}^m \) be the feedforward neural network characterizing an agent's network policy, where,  \(\pmb{x} \in \mathbb{R}^n\)  is the input image to the network, and \(f(\pmb{x})\in \mathbb{R}^m\) is its output. The function \(f\) representing the neural network can be defined as \(f(\pmb{x})=f_L \circ f_{L-1} \circ ... \circ f_1(\pmb{x})\) with each \(f_{l\in L}\) representing a transformation at layer \(l\).

Applied to an input image \(\pmb{x}\), LRP \cite{bach2015pixel,lapuschkin2019unmasking} calculates a relevance score \(R_p(\pmb{x})\)  for each pixel \(p \in \pmb{x}\), indicating the importance of the pixel in the network's decision. The output of the LRP method is a relevance map or heatmap, represented as \(\pmb{R}(\pmb{x}) = \{R_p(\pmb{x})\}_{p \in \pmb{x}}\). \(\pmb{R}(\pmb{x})\)   is derived by iteratively backpropagating the network output \(f(\pmb{x})\) from layer \(l+1\) to layer \(l\) according to some backpropagation rules. These backpropagation rules are guided by a relevance model, as detailed in \cite{montavon2017explaining} which can be expressed in their general form by:
\begin{align}
R_{i\in l} = \sum_{j\in l+1} \frac{q_{ij}}{\sum_{i'\in l} q_{i'j}} R_j
\end{align}

where \(R_{i}\) is the relevance of the neuron \(i \in l\) and \(R_{j}\) is the relevance of the neuron \(j \in l+1\).  Here,  \(q_{ij}\) varies depending on the chosen relevance propagation rule, which is contingent on the input domain of the data \cite{montavon2017explaining}. In our experiments, the data consist of pixel values \(p\) or outputs from a ReLU activation function, both of which are positive real numbers. Consequently, we employ the \(z^+\)-rule for backpropagation. Within this framework, \(q_{ij} = x_iw_{ij}^+ \), where  \(x_i\) represents the activity output of neuron \(i\), and \(w_{ij}^+\) is the positive part of the weight connecting neurons \(i\) and \(j\).

\paragraph{Common procedure}

In our research, we want to determine the extent to which specific regions of an input image \pmb{x} contribute to the activation of a particular neuron in the $F_c$ layer. Given input \pmb{x}, we consider where a neuron is looking at in the input. To do so we apply the following steps for each input, \(\pmb{x} \in \pmb{X}\):
\begin{enumerate}
    \item Calculate the distributed relevance scores \(R_{F_c} = \{R_i\}_{i \in F_c}\). Identify and select the subset of neurons \(S \subseteq F_c\) that collectively account for 90\% of the total relevance \(R_{F_c}\).
    \item For each neuron \(k \in S\), generate the corresponding relevance map in the input space denoted \(\pmb{R}^k(\pmb{x}) = \{R_p^k(\pmb{x})\}\).
    \item Identify the objects highlighted by the relevance maps.
\end{enumerate}

\subparagraph{1. Calculate the distributed relevance scores \(\{R_i\}_{i \in F_c}\)}\mbox{}\\
To compute \(\{R_i\}_{i \in F_c}\) we backpropagate the output relevance \(\pmb{R}_{\text{output}}\) back to the neurons in the \(F_c\) layer using the propagation formula applied with the \(z^+\)-rule. We initialize \(\pmb{R}_{\text{output}} = f(\pmb{x})\) and then following equation (1): 
\begin{align}
R_{i\in F_c} &= \sum_{j\in f(x)} \frac{x_iw_{ij}^+}{\sum_{i'\in F_c} x_i'w_{i'j}^+} \cdot j
\end{align}

We then order the neurons according to their relevance score and select the smallest subset $S$ responsible for at least 90\% of the total relevance score.

\subparagraph{2. For each neuron \(k \in S\) generate \(\pmb{R}^k(\pmb{x}) = \{R_p^k(\pmb{x})\}_{p \in \pmb{x}}\)}\mbox{}\\
Here \(R^k(\pmb{x})\) is the relevance of neuron \(k\) in layer $F_c\) with regard to the network's output \(f(\pmb{x})\). \(R_p^k(\pmb{x})\) is the relevance of the pixel \(p\in x\) to the output of the neuron \(k\).

To compute \(\{R^k(\pmb{x})\}_{k\in S}\) we backpropagate the relevance score of neuron \(k\) to the input space. In this step, we are only interested in \textit{where} the neuron \(k\) is looking on the input image and not its relative contribution to the choice of the action. We therefore initialise the new relevance distribution in the \(F_c$ layer by setting \(R_{F_c} = (t_1,..,t_k,..,t_N)\) where \(t_i=0\) for all \(i \neq k\) and \(t_k=1\) where \(k\) is the index of the neuron under study.
The relevance \(R_{F_c}\) is then backpropagated through the layers of the network until the input \(\pmb{x}\) using the backpropagation formula in (1), such that,
\begin{align}
R_i = \sum_j \frac{x_iw_{ij}^+}{\sum_{i'} x_i'w_{i'j}^+} R_j
\end{align}

This results in a relevance map over the input \(\pmb{x}\) which in combination with the corresponding labeled input \(\bar{\pmb{x}}\) allow us to retrieve where the input the network is looking. We used  \cite{blogpost} for the implementation of the LRP method.

\begin{figure}
    \centering
    \includegraphics[width=0.7\linewidth]{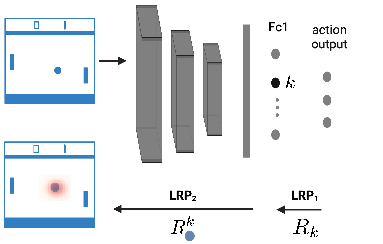}
    \caption{Illustration of the extraction of the relevance score for the ball object computed from a neuron \(k\) in the \(Fc\) layer. Here a single frame (among the 4 frames constituting an input) is represented for clarity purposes. The relevance score of neuron \(k\) with respect to the output of the network is computed with a first LRP operation noted here as \(LRP_1\). The relevance score of the ball object with respect to the neuron \(k\) is then computed with a second LRP operation noted here as \(LRP_2\). }
    \label{fig:LRP-illustration}
\end{figure}

\section{RL training procedure}\label{RL-training-procedure}

For all games, the learning rate was set to \(\alpha = 7e-4\), the discount factor \(\gamma = 0.99\), the entropy coefficient \(\tau = 0.01\) and the value loss coefficient \( v_{lc}= 0.25\). The number of parallel environment was set to \(n = 100\). Training was carried out on Nvidia via A100 GPUs using a single GPU with up to 18 CPU cores per task and a memory of 125 GB.

\section{Game}\label{Game}

\subsection{Game Dynamics Setup}

We developed a modified version of the classic Atari Pong game using the Pygame library (v.2.5.2 \cite{pygame}) to create a custom environment for our experiments.  This variation maintains the original game’s elements—paddles, walls, and scoring—but introduces dual balls instead of one. The dimensions and colours of the game components mirror those in the original Pong. In this setup, each paddle is distinctively colored, as are the two balls. The paddles move on the y-axis at a rate of 2 pixels per frame, while the balls move at speeds of 4 pixels per frame along the x-axis and 2 pixels per frame on the y-axis.Game states are represented as 84x84 pixel RGB images.


\subsection{Preprocessing and Environment Wrapping}

To prepare the game environment for reinforcement learning (RL), we encapsulated it within the same preprocessing wrapper used for Atari games in the Stable Baselines 3 framework (v.3 \cite{stable-baselines3}). This wrapper converts each game frame from RGB to grayscale to streamline input dimensions and reduce computational demands \cite{mnih2013playing}. To prevent the RL agent from memorising action sequences, we introduced randomness in the initial ball direction. Last but not the least, the observation input given to the agent is not Markovian as all the information necessary to predict the next input given this input and an action is not available.  Aligning  with prior research by Y and common practices in Stable Baselines 3, we stacked four consecutive frames, producing a composite input  \(o_t \in R^{4\times84\times84}\), to provide the agent with a temporal context for decision-making.

\subsection{Variations}
The game variations are summarised in Table Y.

\begin{table}
\caption{Design of the game variations.}
\label{tab:game_version}
\centering
\begin{tabular}{lllllll}
Version & \multicolumn{2}{l}{Balls} & \multicolumn{2}{l}{Dynamics} & \multicolumn{2}{l}{Rewards} \\ \hline
        & B1          & B2          & B1            & B2           & B1           & B2           \\ \cline{2-7} 
v0      & Yes         & No          & D1            & X            & R1           & X            \\
v1      & Yes         & Yes         & D1            & D1           & R1           & X            \\
v2      & Yes         & Yes         & D1            & D2           & R1           & R2           \\ \hline
\end{tabular}
\end{table}

\textbf{Dynamics:}
\begin{itemize}
    \item D1: The ball rebounds off both the walls and the paddles.
    \item D2: The ball is capable of bouncing off the walls and the agent’s paddle. However, it will not rebound off the opponent’s paddle, and pass through it. 
\end{itemize}

\textbf{Rewards:}
\begin{itemize}
    \item R1: The agent is awarded +1 for scoring a point on the opponent’s side, and receives a -1 if the agent fails to hit the ball, allowing it to pass by.
    \item R2: The agent gains +1 for successfully hitting the ball with their paddle. Conversely, a -1 penalty is applied if the agent fails to hit the ball, allowing it to pass by.

\end{itemize}

The balls start in the middle of the screen. They have the same x-direction that is randomly generated and opposite y-directions. This choice of initial states forces the agent to choose between the balls for the first hit. The condition for an episode to end and for the balls to respawn is dependent on \textit{B1} being scored. The opponent is hard-coded to position its y-axis on the y-axis of \textit{B1}.

\section{Example combinatorial-attention}\label{combinatorial_attention_example}

\begin{figure}
    \centering
    \includegraphics[width=0.75\linewidth]{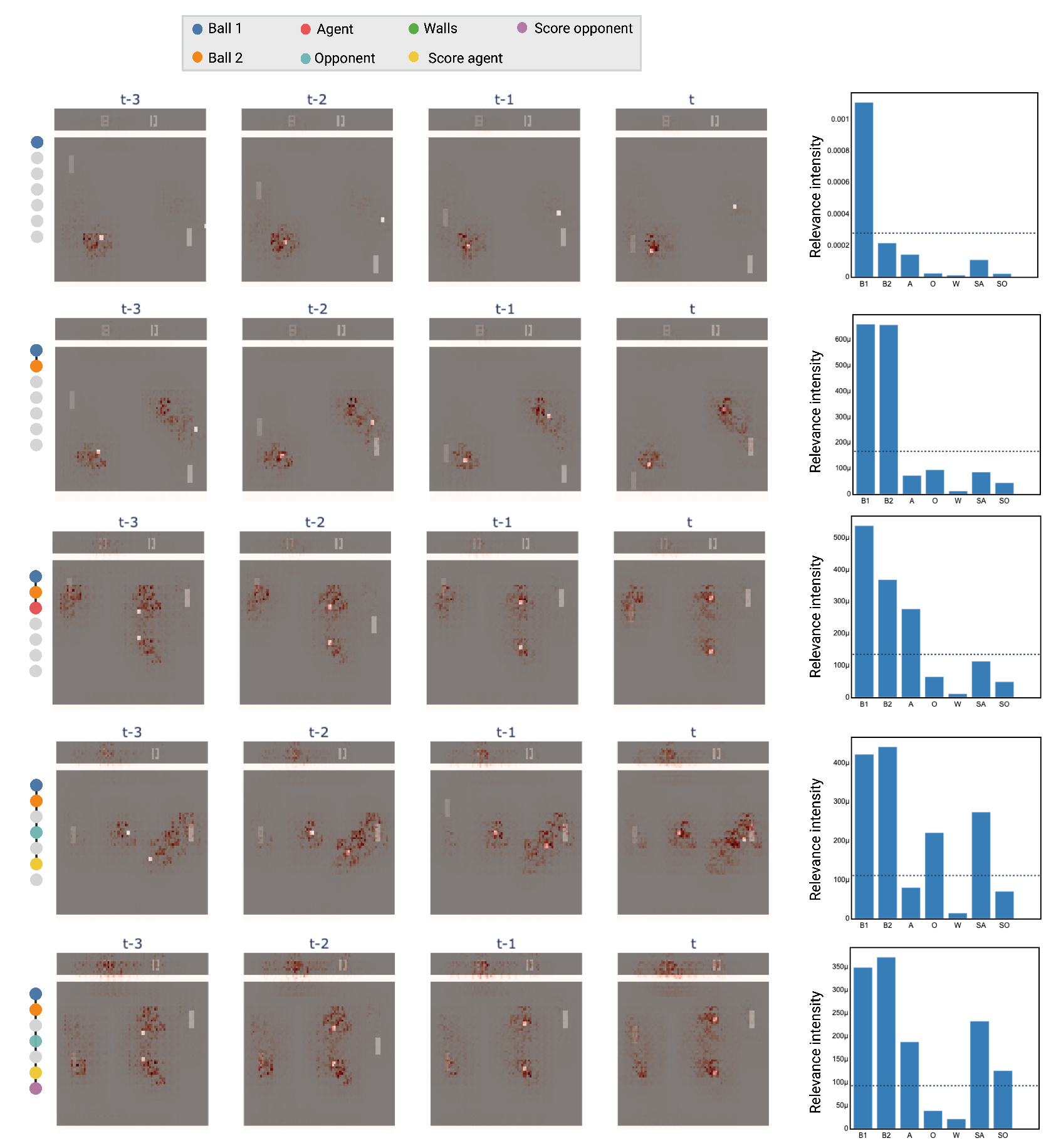}
    \caption{Illustrations of different combinatorial-attention results. Each line represents a combination of objects "looked at" by a neuron in the $F_c$ layer. From left to right: symbol of the combination as depicted in the main text, neuron's relevance map overlaid onto the input frames, bar plot quantifying the average intensity of relevance scores for each object with the 25\% threshold.}
    \label{fig:example_combin_attention}
\end{figure}

Figure \ref{fig:example_combin_attention} presents various combinatorial-attention patterns for individual neurons in the $F_c$ layer. Each neuron's relevance map, generated for a specific input, is overlaid onto the input provided to the agent. The right side features a bar chart that quantifyes the average intensity of relevance scores for each object, with a dotted line indicating the 25\% threshold of the maximum intensity value.

\section{Threshold value for the combinatorial-attention}\label{threhold_value_combinatorial_attention}

\begin{figure}
    \centering
    \includegraphics[width=0.75\linewidth]{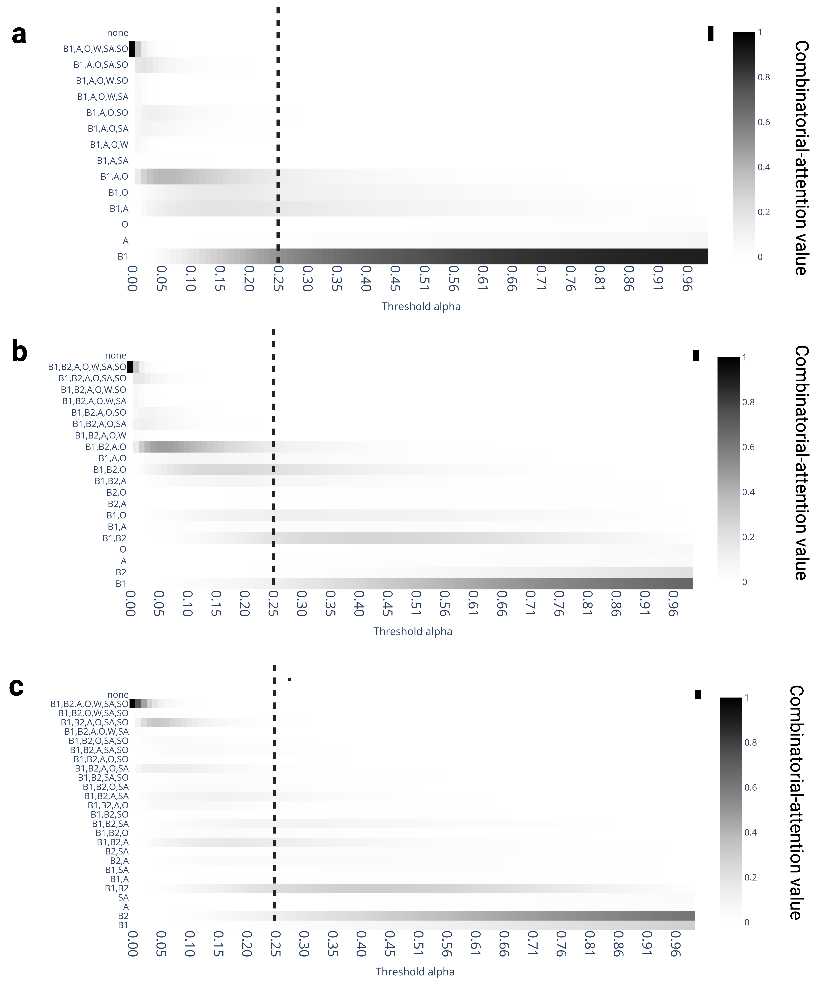}
    \caption{Combinatorial-attention values in function of alpha averaged over 20 trained agents for \textbf{a} version \textit{v0},\textbf{b} version \textit{v1} and \textbf{c} version \textit{v2}. Dotted lines indicate $\alpha = 0.25$.}
    \label{fig:threshold_value}
\end{figure}

The threshold value, $\alpha$, utilised in the combinatorial-attention metric functions as a hyperparameter and influences the result of the metric. This threshold sets the minimum intensity level necessary for an object to be considered within the set of observed objects. A threshold approaching zero will result in a combinatorial-attention metric that accounts for all objects with any non-zero intensity score. Conversely, a threshold near one will reflect combinatorial-attention equivalent to only the single object with the highest intensity value. To avoid these extremes, we choose a threshold of $\alpha = 0.25$. This value allows for the detection of more nuanced patterns of combinatorial attention. To demonstrate the dependency of the combinatorial-attention metric on the threshold value, calculations were performed across a range of threshold values from 0 to 1 for the 20 trained agents in each version of the game. Figure \ref{fig:threshold_value} illustrates the average combinatorial attention metric between models as a function of $\alpha$ for each version of the game.

\section{Impact of the agent and opponent score displayed on the agents' behavior } \label{impact_score_choice_v2}

\begin{figure}
    \centering
    \includegraphics[width=0.8\linewidth]{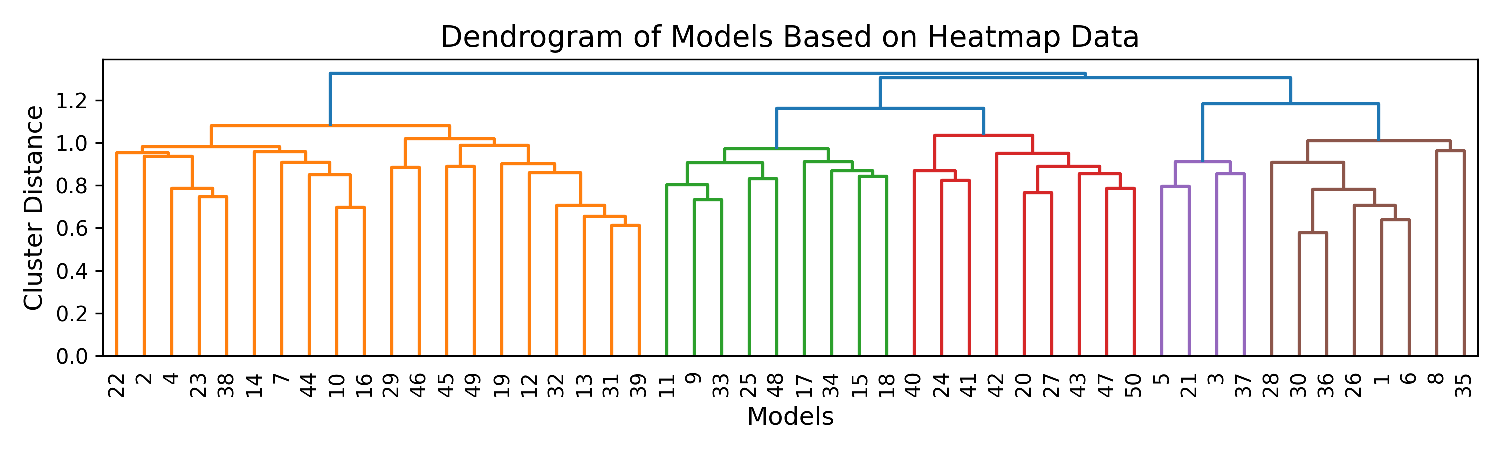}
    \caption{Dendrogram computed from the standardized interaction matrices of 50 agents trained on \textit{V2}. Each interaction matrix is of size 21x41 corresponding to all combinations of score displayed.}
    \label{fig:dendrogram}
\end{figure}

\begin{figure}
    \centering
    \includegraphics[width=0.8\linewidth]{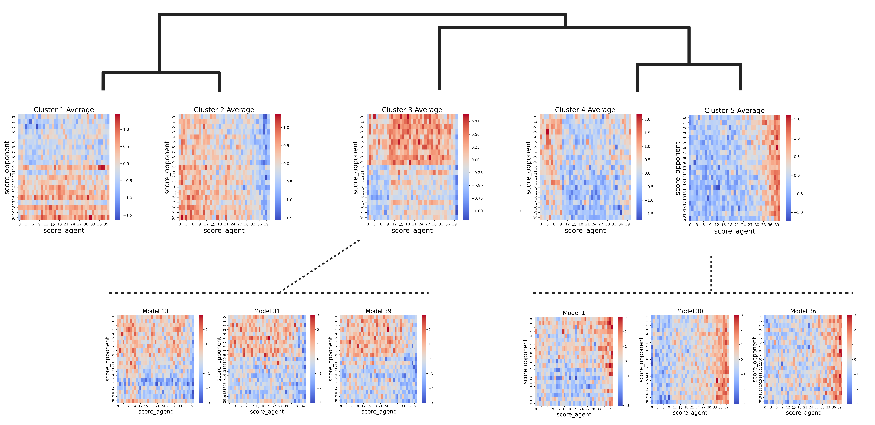}
    \caption{Dendrogram illustrating the hierarchical clustering of agents, with clusters identified up to a distance of 1.1, resulting in five main clusters. The heatmaps for each cluster were generated by averaging the heatmaps of all models within that cluster. The second row presents individual heatmaps from representative agents within clusters three and five.}
    \label{fig:heatmap}
\end{figure}

Based on the observation that ATOMs exhibited a stronger focus on the scores of both the agents and their opponents in version \textit{v2} compared to versions \textit{v1} and \textit{v0}, we investigated the potential impact of the displayed scores on the agent's decision to hit \textit{B1} or \textit{B2}.To examine this, we conducted the \textit{Dual Ball Discrimination Test} across all combinations of displayed scores (ranging from 0 to 20 for the opponent and 0 to 40 for the agent) for each of the 50 agents trained on \textit{v2}. In this test, the agent, given a specific score combination (score agent, score opponent), had to choose to hit \textit{B1} or \textit{B2} over 100 different trajectories. The set of 100 trajectories was consistent across all score combinations and agents. This process generated two-dimensional matrices of dimensions 21 x 41, where each matrix value represents the relative interaction of the agent with \textit{B1}, calculated as the ratio of the number of times the agent hit \textit{B1} to the total number of times the agent hit any ball. We standardized the relative interaction  for each agent to account for variations in the average interactions of different agents. Using the standardized data, we computed a pairwise distance matrix with the 'correlation' metric and performed hierarchical clustering using the complete method. The resulting dendrogram is presented in Figure \ref{fig:dendrogram}.
Figure \ref{fig:heatmap} illustrates the averaged heatmaps by clusters at a distance of 1.1, along with examples of heatmaps for models belonging to clusters three and five. Positive values (in red) indicate a preference for \textit{B1}, while negative values (in blue) indicate a preference for \textit{B2}. The findings reveal that the displayed scores influence the agent's choice to hit \textit{B1} or \textit{B2}, with distinct patterns of influence emerging among different agents. For example, agents belonging to cluster five mostly interact with \textit{B1} until their displayed score get closer to 40 points where they tends to choose \textit{B2} over \textit{B1} (the game terminates when the agent reaches 41 points). In contrast, agents belonging to cluster three interact more with \textit{B1} or \textit{B2} in function of if the score of the opponent displayed is below or above 10 points.

\section{Color swap between B1 and B2}
\begin{figure}
    \centering
    \includegraphics[width=0.75\linewidth]{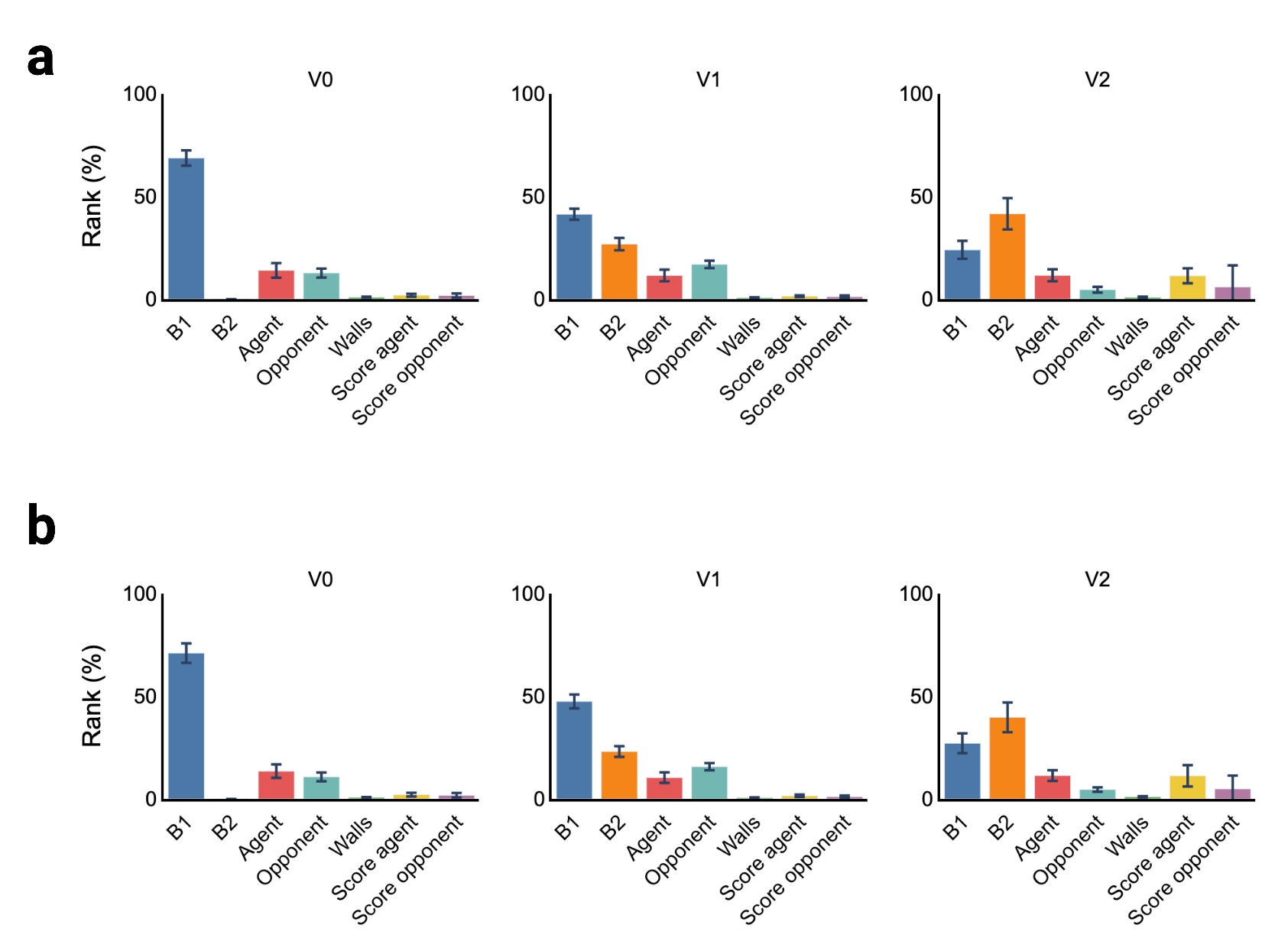}
    \caption{\textbf{a} Hierarchical attention average across 47 agents with an initial color mapping: B1 color (236, 236, 236) and B2 color (255, 255, 0). \textbf{b} Hierarchical attention average across 20 agents with a swapped color mapping: B1 color (255, 255, 0) and B2 color (236, 236, 236). Error bars represent the standard deviation.}
    \label{fig:color_swap}
\end{figure}

In this experiment, we investigated whether the colour assigned to each ball affected the outcomes as measured by our metrics. We conducted this experiment by training 20 models for each game version, exchanging only the colors of balls B1 and B2. As observed on Figure \ref{fig:color_swap}, the color of the ball does not impact the hierarchical-attention pattern which similar for both mapping of color to ball.

\section{ATOMs for version v2 complete}\label{hierarchical} 


\begin{figure}
    \centering
    \includegraphics[width=0.8\linewidth]{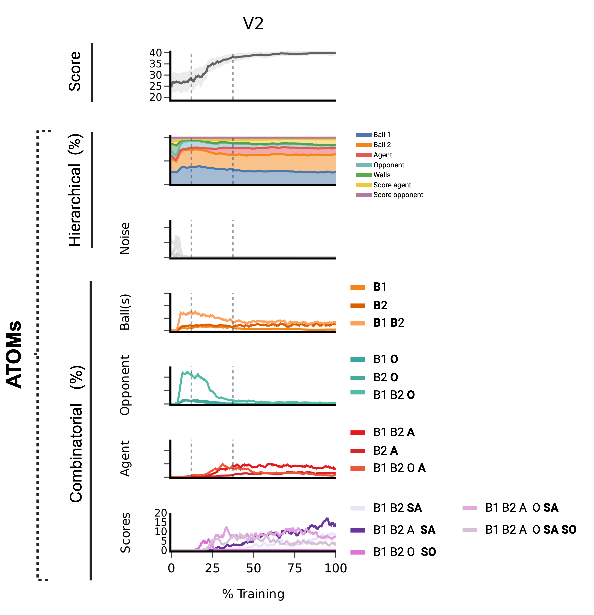}
    \caption{Development of the performance score and of ATOMs during learning for an agent trained on version \textit{v2}. The performance scores of the agent is computed at each measurement over ten games. The dotted lines indicate the time frame during which the agent's score showed a marked increase. These time frames were estimated manually.The average score is depicted by the central dark line, with the shaded region representing the standard deviation. The combinatorial-attention is categorized as follows from top to bottom: noise (grey), combinations of balls only (orange), combinations of opponent and balls (cyan),combinations of objects including the agent itself (red), and combinations of objects including the score opponent (SO) and the score of the agent (SA) (purple). }
    \label{fig:learning_v2}
\end{figure}

\end{document}